# Fast Hierarchical depth map computation from stereo

Vinay Kaushik, Brejesh Lall
Dept. of Electrical Engineering
IIT Delhi, India
India
{eez158117, brejesh}@ee.iitd.ac.in

*Abstract*—Disparity by Block-Matching stereo is usually used in applications with limited computational power in order to get depth estimates. However, the research on simple stereo methods has been lesser than the energy-based counterparts which promise a better quality depth map with more potential for future improvements. Semi-global-matching (SGM) methods offer good performance and easy implementation but suffer from the problem of very high memory footprint because it's working on the full disparity space image. On the other hand, Block-matching stereo needs much less memory. In this paper, we introduce a novel multi-scale-hierarchical-block-matching approach using a pyramidal variant of depth and cost functions which drastically improves the results of standard block matching stereo techniques while preserving the low-memory footprint and further reducing the complexity of standard block matching. We tested our new multi-block-matching scheme on the Middlebury stereo benchmark. For the Middlebury benchmark we get results that are only slightly worse than state-of-the-art SGM implementations.

I. INTRODUCTION

Estimating depth information from a stereo-camera is still one of the most versatile solutions for 3-D sensing with a wide application range in robotics[1][2], intelligent vehicles and also space science. The drawback of estimating depth with a stereo camera system, with respect to direct systems like LIDAR or time-of-flight, is the comparably high computational effort which is necessary to extract the depth Information from the stereo images by finding correspondences. On the benefit side stereo cameras provide depth information with a very high spatial resolution which is a prerequisite for obstacle avoidance or path planning. Furthermore, the images provided by the cameras allow for other usage like object recognition or ego-motion estimation. In the literature there are two main categories of algorithms for finding the stereo correspondences: local and global methods. Local methods typically find correspondences by matching patches of one stereo image to the other image. In contrast, global methods typically optimize for an energy function that describes the best transformation of one image into the other[3]. Usually this involves some smoothness or regularization terms in order to tackle NP-completeness for feasible processing. Apart from these two large groups there is one method that is located in between: semi-global-matching (SGM)[4]. On the one hand SGM is also based on an energy-functional, on the other hand it does not employ a fully global optimization but optimizations along one-dimensional paths. This semi-global optimization scheme has a depth accuracy that comes close to global stereo methods but with a much lower computational complexity. Due to this, SGM has become very popular, especially in the domain of intelligent vehicles. One major drawback of SGM is its high memory footprint because it requires the full disparity space image (DSI). This property makes it very challenging to bring SGM to low-energy hardware means like FPGA.

In contrast, local stereo methods based on block-matching are very easy to port to various hardware architecture because they need only a small part of the DSI at a time and the processing is embarrassingly parallel. The downside of local methods is a generally lower accuracy and density of the resulting depth maps. In this paper, we will introduce a novel multi-scale-hierarchical-block-matching (MSIBM) scheme that uses the smallest scale Gaussian stereo pair to compute the reduced depth map and uses the depth at that lower scale to optimally compute the depth at higher scale. This goes on hierarchically to get the final depth.

This scheme leads to a significant improvement in both speed and accuracy over standard block-matching (BM) stereo while still preserving the low-memory and high-parallelization properties. Moreover, it is more robust to image noise. Our experiments with this novel block-matching scheme on the Middlebury stereo benchmark show a major improvement with respect to standard BM stereo.

II. PROBLEMS IN STEREO BLOCK-MATCHING

The basic idea of standard block-matching (BM) stereo is very simple. By correlating image patches (called blocks or filters) between the left and right stereo images, correspondences between the images are found. The position difference of a correspondence is called disparity and is inversely coupled to the distance. There is a large bunch of cost functions [10] used as matching criteria; however, typically sum of absolute difference (SAD), normalized cross-correlation (NCC), rank transform (RT)[5], census transform (CT)[6] zero-mean normalized cross-correlation (ZNCC)[7] are used.

Disparity is defined as the distance between similar points in 2 images. To find those similar points various standard block matching approaches have been defined. The similarity is defined on some mathematical score such as correlation, sum of absolute difference (SAD), sum of squared difference (SSD), NCC and ZNCC. For a NxN image having maximum disparity D, the disparity of a point is computed by comparing the pixel in right image to a pixel in the left image (on the same epipolar line) D number of times. Considering a block of size MxM, the complexity of the program to compute disparity map is an order of 'D'. There are several disadvantages of standard block matching techniques. Computing depth maps on higher resolution images not only takes time, but also creates artefacts. Little noise can lead to a very poor disparity map. Many post processing steps (filters) such as WLS filter and anisotropic median filter are applied to improve the generated map. This post processing is not possible in real time applications. Not only it takes time but excessive global filtering also leads to loss of useful depth information.

We propose a novel depth map computation mechanism using a Stereo-Gaussian pyramid that eliminates the requirement of post-processing (filtering) of depth images, takes less time and creates maps for different spatial resolution that are more accurate than standard block matching methods.

We compute depth maps from stereo images that are accurate as well as take minimal time as compared to other algorithms. Using parallelization at its root level helps in performing optimizations in parallel and saves crucial time. We use a Stereo-Gaussian pyramidal variant to compute depth maps of multiple spatial resolution and hierarchically fuse them to compute a map that is more accurate and needs less computations than standard block matching scheme.

III. APPROACH

A standard block matching algorithm uses a cost function for matching image patches along epipolar lines to find the best match, thus defining the disparity of the point. For every pixel, the algorithm searches from 0 to $d_{max}$, where $d_{max}$ is the maximum disparity level for that stereo pair. The computation is not only time consuming, but the disparity filtering is done as a post processing step to the disparity image. There have been various ways which reduce the disparity search along the epipolar lines, but all such methods consider certain probabilistic assumptions and are themselves an overhead for the system. We define a novel algorithm which drastically reduces the number of comparisons required to compute the disparity for a pixel.

For a stereo image pair, we first define a Depth Search Image (DSI) as follows:

$$E_{\{i,j\}}^k(z) = f_{ZNCC}\left(L_{\{i,j\}}, R_{\{i,j+z\}}\right) \qquad (1)$$

This defines the matching cost for pixel *(i,j)* in left image at disparity z, where $z \in \{N^+; 0 \leq z \leq d_{max}\}$ and $f_{ZNCC}$ computes a zero-mean normalized cross-correlation between patches L and R.

The ZNCC is defined as: $ZNCC(I_1, I_2, u_1, v_1, u_2, v_2, n) :=$

$$\frac{\frac{1}{2n+1}\sum_{i=-n}^{i=n}\sum_{j=-n}^{j=n}\pi_{t=1}^{2}\left(I_t(u_t+i,v_t+j) - \bar{I}(u_t,v_t,n)\right)}{\sigma_1(u_1,v_1,n).\sigma_2(u_2,v_2,n)} \qquad (2)$$

Where, $I_1$ and $I_2$ are stereo images and the patches $L$ and $R$ are of size $2n + 1$ with $(u_i, v_i)$ as patch centres.

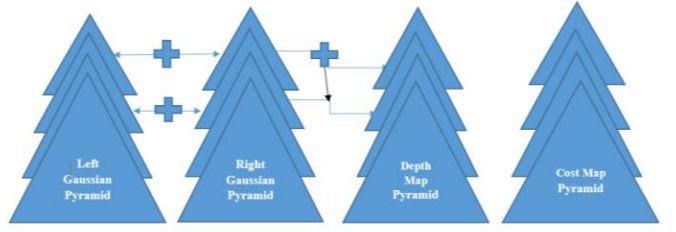

Fig. 1. Block Diagram of Pyramidal Disparity Map Computation Method

We construct a K-level Gaussian pyramid for both left and right pair of stereo images. The maximum disparity at $k^{th}$ level is defined as $d_{max}^k = \frac{D_{max}}{2^k}$, where $D_{max}$ is the maximum disparity level for the original image.

The matching block size for kth level decreases by a factor of 2 at every level of Gaussian pyramid. The disparity map computation is further divided into 2 parts. Fig. 1 gives the block diagram of the disparity map computation scheme.

Taking the highest level image, we perform standard block matching at that scale to find its disparity map. We further refine the map by performing optimization of pixels with poor matching cost.

A. *Selecting Disparity for highest Gaussian level stereo pair*

Taking the highest level image, we perform standard block matching at that scale to find its disparity map. We further refine the map by performing optimization of pixels with poor matching cost.

We compute disparity as the best cost match from the Disparity Search Image. Thus,

$$d_{i,j}^{\prime k} = \underset{z \in (0, d_{max}^k)}{\arg\max} E_{i,j}^k(z), \text{ and} \qquad (3)$$

$$C_{i,j}^{\prime k} = \underset{z \in (0, d_{max}^k)}{\max} \{E_{i,j}^k(z)\} \qquad (4)$$

Here, $d_{i,j}^{\prime k}$ is the calculated disparity for the pixel$(i,j)$ and $C_{i,j}^{\prime k}$ is the matched cost for pixel$(i,j)$.

B. *Disparity Refinement*

For disparity refinement, we compute an average DSI for every pixel with low matching cost and select the disparity with maximal average matching cost, thus considering disparity as correct if multiple disparity searches around the neighborhood of the point lead to same disparity. Thus,

$$\widehat{E_{i,j}^k}(z) = \sum_{(m,n)\in N(i,j)} E_{m,n}^k(z) \qquad (5)$$

Here, $\widehat{E_{i,j}^k}(z)$ is the average DSI for the pixel$(i,j)$.
The final disparity at level $k$ is given by:

$$d_{i,j}^k = \begin{cases} d_{i,j}^{\prime k} & , if\ C_{i,j}^{\prime k} > \alpha \\ \underset{z \in (0, d_{max})}{\arg\max} \widehat{E_{i,j}^k}(z) & else \end{cases} \qquad (6)$$

Similarly, the matched cost at level $k$ is given by:

$$C_{i,j}^k = \begin{cases} C_{i,j}^{\prime k} & if\ C_{i,j}^{\prime k} > \alpha \\ \underset{z \in (0, d_{max})}{\max} \widehat{E_{i,j}^k}(z) & otherwise \end{cases} \qquad (7)$$

Here, $d_{i,j}^k$ is the final disparity for the pixel$(i, j)$ and $C_{i,j}^{\prime k}$ is the matched cost for pixel$(i, j)$.

## C. Hierarchical Disparity Computation

We then upsample the smaller depth map as well as the cost map using nearest neighbour and bi-cubic interpolation respectively as interpolation method to generate a depth and cost map for lower Gaussian level of stereo pair. Thus, $\widehat{d_{i,j}^k} = NN(d_{i',j'}^{k+1})$ and $\widehat{C_{i,j}^k} = BI(C_{i',j'}^{k+1})$ are the interpolated depth and cost values at $k^{th}$ level. Here, NN is the nearest neighbor interpolation method and BI is the Bi-cubic interpolation method.

This disparity image is not an optimal one but acts like a prior for computing the disparity map at this level. Depending on the interpolated cost, we divide the disparity computation in 2 steps.

$$d_{i,j}^{\prime k} = \begin{cases} \underset{z \in (\widehat{d_{i,j}^k}-1, \widehat{d_{i,j}^k}+1)}{\arg\max} E_{i,j}^k(z) & if\ \widehat{C_{i,j}^k} > \beta \\ \underset{z \in (0, d_{max}^k)}{\arg\max} E_{i,j}^k(z) & otherwise \end{cases} \quad (8)$$

$$C_{i,j}^{\prime k} = \begin{cases} \underset{z \in (\widehat{d_{i,j}^k}-1, \widehat{d_{i,j}^k}+1)}{\max} E_{i,j}^k(z) & if\ \widehat{C_{i,j}^k} > \beta \\ \underset{z \in (0, d_{max}^k)}{\max} E_{i,j}^k(z) & otherwise \end{cases} \quad (9)$$

Here, $d_{i,j}^{\prime k}$ is the computed disparity at level $k$ with matching cost $C_{i,j}^{\prime k}$.

To compute the optimal depth map at $k^{th}$ level, instead of performing comparisons till maximum disparity, we use standard block matching only if the pixel's matching cost is very low (i.e. a bad match in case of ZNCC). For most of the pixels with good matching cost, we perform disparity refinements i.e. we begin matching near the disparity obtained from the interpolated image. The number of comparisons for refinement per pixel depends on the level of the image, and in our case since we halved the image to construct Gaussian pyramid, it takes just 3 comparisons per pixel to find the optimal disparity of the larger stereo pair.

After this step, selective median filtering is applied to every bad pixel to further optimize the disparity map.

## D. Selective Median Filtering

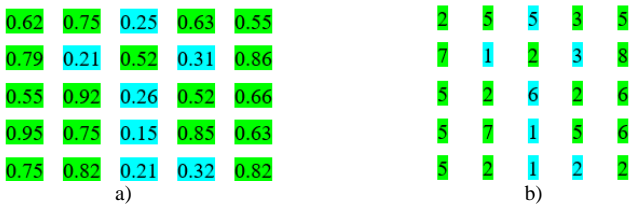

Fig. 2. a) Cost match for a 5x5 patch  b) Disparity Map for a 5x5 patch. Median filtering is performed on selected green colour disparity values.

To further refine the disparity image, we apply a variant of median filtering on the pixels with low matching cost. For every such pixel, we apply a 5x5 patch around it and discard the pixels with bad matching cost from that patch. As shown in the above Fig. 2, we select only those pixels which have good matching coast and find their median and assign it to the centre pixel.

$$d_{i,j}^k = \begin{cases} d_{i,j}^{\prime k} & if\ C_{i,j}^k > \alpha \\ \underset{(m,n) \in N(i,j)}{median}\ d_{i,j}^{\prime k} & otherwise \end{cases} \quad (10)$$

Here $d_{i,j}^k$ is the optimal disparity map of the $k^{th}$ level. We follow this approach hierarchically to compute the disparity map for the original image.

The above approach incorporates the best of information from the lower scale level of Gaussian pyramid and block matching at current scale to compute a disparity map which is more robust as well as faster than the standard block matching mechanisms.

## IV. RESULTS

Middlebury V3 dataset [8] is used for experimentation. It consists of 15 test images. It introduces more challenging scenarios such as high-resolution stereo pair images, varying exposure and lighting settings, imperfect rectification, etc. We compare our technique with several other state of the art BM and SGM methods. We also provide the runtime evaluations on the methods which run on CPU based platforms (Table I).

TABLE I.    AVERAGE RUNTIME EXECUTION

| Method | Environment | Average Runtime (in min) |
|---|---|---|
| BM stereo | MATLAB | 50 |
| BM stereo (multithreaded) | MATLAB | 10 |
| Displets v2 [9] | MATLAB + C/C++ | 4 |
| PSPO [10] | MATLAB + C/C++ | 5 |
| PRSM [11] | C/C++ | 5 |
| ISF [12] | C/C++ | 10 |
| Ours | MATLAB | 2 |

Our method performs resonably well in terms of both speed as well as accuracy. The method is comparable to the best SGM methods and outperforms all BM methods. We evaluate our method on basis of average error computed by the Middlebury Eval 3.0 toolkit (Table II). We use a block size of 11x11 for the lowest level disparity image. The theresholding parameters include $\alpha\ and\ \beta$. If $\alpha, \beta$ tends to 0, although the method takes less time but there is no disparity refinement. If $\alpha, \beta$ tends to 1, it recomputes the disparity of every pixel on the basis of its neighbors which increases computational complexity drastically and softens disparity boundaries. Thus we select $\alpha$ and $\beta$ as 0.9 for our experiments as it gives the optimal result. Since it is at the highest level, it needs less number of comparisons. And since we use a Gaussian blur over original image, the matching is robust to some amount of noise in the image. Fig.3 shows the disparity maps of few example images of the Middlebury V3 dataset.

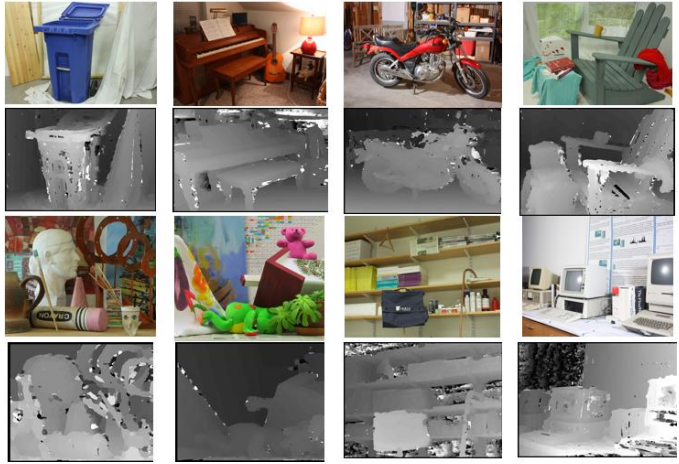

Fig. 3. 1st and 2nd Row: Original Image, 3rd and 4th Row: Disparity Map

TABLE II. AVERAGE DISPARITY ERROR

| | | | |
|---|---|---|---|
| Adirondack\|28.61 | ArtL\|33.21 | Playroom\|37.61 | Shelves\|30.08 |
| MotorcycleE\|36.43 | Piano\|26.79 | Playtable\|33.52 | Teddy\|27.41 |
| Jadeplant\|65.64 | PianoL\|26.79 | PlaytableP\|33.65 | Vintage\|61.66 |
| Motorcycle\|36.43 | Pipes\|31.46 | Recycle\|25.68 | Average\|:35.6 |

## V. CONCLUSION

We have proposed multi scale hierarchical block matching algorithm using pyramidal variant of depth and cost maps. We exhaustively evaluate it against disparity computation techniques which require low processing power and demonstrate that the proposed approach achieves high accuracy while having low computational cost. This can be attributed to the selective cost optimizations using lower scale depth map and cost map as a prior limiting the disparity search space.